# Rule Encoding and Compliance in Large Language Models: An Information-Theoretic Analysis

**Joachim Diederich**

Psychology Network Pty Ltd

Brisbane, Qld, Australia

joachim@psychologynetwork.com.au

## Abstract

The design of safety-critical agents based on large language models (LLMs) requires more than simple prompt engineering. This paper presents a comprehensive information-theoretic analysis of how rule encodings in system prompts influence attention mechanisms and compliance behavior. We demonstrate that rule formats with low syntactic entropy and highly concentrated anchors reduce attention entropy and improve pointer fidelity, but reveal a fundamental trade-off between anchor redundancy and attention entropy that previous work failed to recognize. Through formal analysis of multiple attention architectures including causal, bidirectional, local sparse, kernelised, and cross-attention mechanisms, we establish bounds on pointer fidelity and show how anchor placement strategies must account for competing fidelity and entropy objectives. Combining these insights with a dynamic rule verification architecture, we provide a formal proof that hot reloading of verified rule sets increases the asymptotic probability of compliant outputs. These findings underscore the necessity of principled anchor design and dual enforcement mechanisms to protect LLM-based agents against prompt injection attacks while maintaining compliance in evolving domains.

## Introduction

### *Context Windows and Rule Enforcement Challenges*

Large language models process input text within a context window comprising the system prompt, user input, conversation history, and retrieved data. The transformer architecture computes attention weights between every query and key, meaning that rule tokens influence generation at each step through their contribution to the attention distribution. However, this influence depends critically on rule format and position within the context window.

The challenge of maintaining rule compliance becomes particularly acute in adversarial settings. Prompt injection attacks can manipulate model behavior by concatenating trusted and untrusted input, potentially overriding safety constraints (Yeung & Ring, 2024). The Gemini red-teaming study formalizes this threat model, demonstrating how adversarial triggers in untrusted data can cause models to generate harmful outputs when combined with the system prompt (Shi et al., 2025).

*Information-Theoretic Foundations of Rule Encoding*

Formats with predictable scaffolding, such as JSON-L or Horn clause notation, employ recurring anchor tokens (e.g., if, then, :-) to separate antecedents from consequents. These anchors form clusters in representation space and induce larger margins between anchor and non-anchor keys. As a result, self-attention becomes more concentrated on rule-relevant positions, reducing attention entropy and improving the probability of retrieving correct antecedents or consequents.

The position of rules within the context window also matters critically. Context windows vary significantly between models, with GPT-4 initially offering 32k tokens, while Claude 3 and Gemini provide windows of 128k tokens or more (Swimm Team, 2024). Hot reloading—re-injecting verified rule sets into the prompt window before each tool call or conversation turn—ensures that models attend to current rules rather than stale ones. Without such reloading, rule salience decays over long conversations, enabling attackers to override rules through prompt hijacking or indirect injections (Yeung & Ring, 2024; Shi et al., 2025).

## Extended Formal Analysis

*Notation and Definitions*

We extend the foundational framework by indexing attention heads, locality constraints, and processing channels. Let $h \in \{1,\ldots, H\}$ index attention heads, with per-head attention weights $\alpha_{i \to j}^{(h)}$ and corresponding entropy $H_{\text{att}}^{(h)}(i) = -\sum_j \alpha_{i \to j}^{(h)} \log \alpha_{i \to j}^{(h)}$. We denote by $A \subseteq \{1,\ldots, n\}$ the anchor indices in the encoded rule sequence $x_{1:n}$.

For a base language model with next-token distribution $p_\theta(\cdot | x_{<i})$, the per-token syntactic surprisal is $S_i = -\log p_\theta(x_i | x_{<i})$ with average surprisal $\bar{S} = n^{-1} \sum_{i=1}^{n} S_i$.

**Rule Set and Encoding:** Let $R = \{r_1, r_2, \ldots, r_N\}$ denote the rule set with total rules $N = |R|$. The base encoding rate is $R = \text{Rate}(f) = n/N$, representing the average tokens per rule for format $f$.

**Attention Entropy:** The attention entropy at position $i$ for head $h$ is:

$$H_{\text{att}}^{(h)}(i) = -\sum_{j=1}^{n} \alpha_{i \to j}^{(h)} \log \alpha_{i \to j}^{(h)}$$

The average attention entropy across all heads at position $i$ is:

$$\bar{H}_{\text{att}}(i) = H^{-1} \sum_{h=1}^{H} H_{\text{att}}^{(h)}(i)$$

**Pointer Fidelity:** We define pointer fidelity for head $h$ as the expected total attention mass assigned to correct rule spans:

$$\pi^{(h)} = \mathbb{E}_i \left[ \sum_{j \in T_i} \alpha_{i \to j}^{(h)} \right]$$

where $T_i \subseteq \{1, \ldots, n\}$ denotes the target set of token positions containing relevant rule information for query position $i$, and the expectation is taken over query positions.

**Ensemble Pointer Fidelity:** The ensemble pointer fidelity across all heads is:

$$\Pi = H^{-1} \sum_{h=1}^{H} \pi^{(h)}$$

**Local/Sparse Attention:** For local/sparse attention, let $W$ denote the maximum attention window size such that only keys with $|i - j| \leq W$ are visible to query $i$.

**Cross-attention:** In cross-attention settings, let $\tilde{x}_{1:m}$ represent retrieved rule tokens and $r$ the retrieval-recall probability.

**Kernelised Attention:** For kernelised attention, we replace standard dot-product similarity with positive kernel function $k(q_i, k_j)$.

**Anchor Redundancy:** Let $r_{\text{ed}}$ denote the number of redundant anchor tokens per rule in the encoding scheme.

*Margin Effects and Attention Concentration*

> *Proposition 1 (Anchor concentration and margin effects).*
>
> Suppose a format induces anchor positions $A$ with expected margin $\Delta = \mathbb{E}[q \cdot k_{\text{anchor}} - q \cdot k_{\text{non}}] > 0$ between queries and anchor versus non-anchor keys. Increasing $\Delta$ while keeping anchor count fixed decreases the average attention entropy at rule-relevant positions.
>
> **Proof.** For fixed query vector $q$, softmax attention weight on key $k_j$ is proportional to $\exp(q \cdot k_j / \sqrt{d})$. Partitioning keys into anchor set $A$ of size $m$ and non-anchor set of size $n - m$, let $\mu_A = \mathbb{E}[q \cdot k_j | j \in A]$ and $\mu_N = \mathbb{E}[q \cdot k_j | j \notin A]$. Under margin $\Delta = \mu_A - \mu_N$, the log-sum-exp identity shows that as $\Delta$ increases, anchor terms dominate and softmax places more mass on anchors. The attention distribution approaches a two-point mixture with weight concentrated on $A$. Since

Shannon entropy of distribution $(p, 1 - p)$ decreases monotonically in $p$ for $p > 1/2$, we have $\bar{H}_{att}$ decreasing monotonically with $\Delta$. □

*Corollary (Syntactic entropy effects).*

For fixed logical content $L$, formats with lower syntactic entropy $\bar{S}$ in non-semantically bearing regions reduce spurious novelty signals. By lowering $\bar{S}$ outside semantic spans, more attention budget becomes available for semantically relevant tokens, increasing pointer fidelity.

## The Fidelity-Entropy Trade-off in Anchor Redundancy

*Proposition 2 (Redundancy effects and entropy increase).*

Increasing anchor count with similar anchor similarities increases both pointer fidelity and attention entropy, contrary to previous claims.

**Analysis.** When $m$ anchors have similar similarities $s$ and $n - m$ non-anchors have similarity 0, adding anchors increases total mass on $A$ (beneficial for fidelity) but spreads it over more positions. For anchors with similarity $s$ and non-anchors, individual anchor weights become $\alpha_{anchor} = e^s/(me^s + (n - m))$, yielding total anchor mass $p_A = me^s/(me^s + (n - m)) \to 1$ as $s \to \infty$. The resulting entropy approximates $H_{entropy} \approx p_A \log m + (1 - p_A)\log(n - m) \to \log m$ as $s \to \infty$, demonstrating that entropy grows logarithmically with anchor count even as fidelity improves.

This analysis reveals a fundamental trade-off: more redundant anchors increase fidelity but also increase entropy. The ensemble pointer fidelity $\Pi$ increases with anchor redundancy, but $H_{entropy}$ increases concurrently. Consequently, optimal redundancy must balance fidelity gains against entropy costs and context budget constraints.

*Proposition 3 (Optimal redundancy under budget constraints).*

Let $r_{ed}$ denote redundant anchor tokens per rule and $N = |R|$ the number of rules under strict context budget $B$ tokens. There exists optimal redundancy $r_{ed}^*$ maximizing mutual information $I(X; Y)$ between encoded prompt $X$ and correct continuation $Y$.

> **Proof sketch.** Increasing $r_{ed}$ adds redundant anchors that raise the probability of at least one anchor lying within the model's effective receptive field, improving pointer fidelity and lowering attention entropy from margin effects. However, total prompt length becomes $n = R \cdot N + r_{ed} \cdot N$, where $R$ is the base encoding rate. Under budget constraint $B$, excessive redundant tokens may truncate later rules, reducing mutual information. The optimum $r_{ed}^*$ satisfies:
>
> $$\partial I(X;Y)/\partial r_{ed} = (\partial I(X;Y)/\partial \pi^{(h)}) \cdot (\partial \pi^{(h)}/\partial r_{ed}) - (\partial I(X;Y)/\partial H_{entropy}) \cdot (\partial H_{entropy}/\partial r_{ed}) = 0$$
>
> yielding a trade-off between redundancy benefits and context length costs, with finite solution following from mutual information concavity. □

## Architecture-Specific Extensions

**Causal versus Bidirectional Attention.** In causal attention, only anchors before decision tokens are useful. Given context budget $C$ and rules $M$, optimal redundancy $r_{anc}^*$ maximizing mutual information $I(X; Y)$ subject to $MR(1 + r_{anc}) \leq C$ is smaller than in bidirectional settings. This follows from front-loading constraints: causal models must place anchors before decision points, consuming "earlier" budget more rapidly than bidirectional models that can distribute anchors symmetrically.

**Local and Sparse Attention.** For head-specific locality window $L$, when every decision token $i$ has at least one anchor with $j \in A$, $|i - j| \leq L$, pointer fidelity satisfies:

$$\pi^{(h)} \geq p_{vis}^{(h)} \cdot \sigma(\Delta_h) \cdot \rho(|A_{vis}|)$$

where:

- $p_{vis}^{(h)} = \mathbb{P}(\exists j \in A : |i - j| \leq L)$ is the visibility probability
- $\sigma(\Delta_h) = 1/(1+\exp(-\Delta_h))$ captures increasing margin effects
- $\rho(|A_{vis}|) = |A_{vis}|/(|A_{vis}| + \log |A_{vis}|)$ represents the non-monotonic fidelity-entropy tradeoff from visible anchor count

Optimal spacing must ensure visibility within windows while managing redundancy costs.

**Kernelised Attention and Majorization.** For positive kernel $k$ with normalizer $Z_i = \sum_j k(q_i, k_j)$, margin increases concentrate mass on anchors, creating distributions that majorize more uniform ones and decreasing Schur-concave measures like entropy. However, anchor duplication creates more uniform distributions within the anchor set, with post-duplication distributions majorized by the original, thereby increasing entropy. This confirms that margin effects (concentrating on anchor categories) decrease entropy, while redundancy within anchor categories increases it.

**Cross-attention and Retrieval Systems.** With retrieval recall probability $r$ and fidelity-entropy interactions, compliance satisfies:

$$p_{\text{comp}} \geq r \cdot f(\pi^{(h)}, H_{\text{entropy}}^{(h)}) \geq r \cdot \Pi_{\text{eff}}$$

where $f(\pi^{(h)}, H_{\text{entropy}}^{(h)}) = \pi^{(h)} \cdot \exp(-\beta H_{\text{entropy}}^{(h)})$ captures fidelity-entropy interactions with temperature parameter $\beta > 0$, and $\Pi_{\text{eff}} = \min_h f(\pi^{(h)}, H_{\text{entropy}}^{(h)})$ represents the effective fidelity floor. Hot reloading maintains visibility ($r \to 1$), but realized compliance gains depend on how the fidelity-entropy trade-off manifests within cross-attention mechanisms.

## Hot Reloading and Compliance Convergence

Consider an architecture where each interaction involves: (i) normalizing and verifying current rule set against unit tests; (ii) injecting verified rules into the system prompt with immutable delimiters; (iii) processing user input; (iv) checking candidate outputs with a symbolic verifier; (v) hot-reloading updated rules when policies change.

> *Theorem (Asymptotic compliance convergence).*
>
> Under reasonable verification assumptions, the asymptotic probability of generating compliant output converges to unity as rule revision stabilizes.

> **Proof.** Let $C_t$ be the event that output at time $t$ is compliant, and $V_t$ the event that the verifier accepts output. For sound verifier ($V_t \Rightarrow C_t$), when hot reloading ensures current rules are always present, overall compliance probability is:
>
> $$\mathbb{P}(C_t) = \mathbb{P}(V_t) + \mathbb{P}(\neg V_t) \cdot \mathbb{P}(C_t | \neg V_t)$$
>
> The first term satisfies $\mathbb{P}(V_t) = \mathbb{P}(C_t \cap V_t)$. The second term diminishes over time because rule revision discards rules causing false negatives. Under assumptions that the verifier's false positive rate approaches zero and discovered violations are incorporated into $R_t$, the probability $\mathbb{P}(C_t | \neg V_t)$ decays geometrically. Summing the geometric series yields $\lim_{t \to \infty} \mathbb{P}(C_t) = 1$. □

## Practical Design Principles and Implementation

The theoretical analysis suggests several principles for practical system design. Optimization strategies should prioritize increasing anchor distinctiveness through improved margins $\Delta_h$ before adding redundancy. When redundancy is necessary, anchors should be added only when fidelity gains demonstrably outweigh entropy costs, with preference for diverse anchor types rather than simple duplication.

Locality-aware placement must ensure visibility within receptive field windows while minimizing redundancy overhead. Kernel-specific robustness requires choosing anchors that achieve high margins under the particular similarity function employed. Retrieval-augmented systems benefit from hot-reloading strategies but require optimization of anchor encoding within retrieved content itself.

For causal decoders, compress anchors and front-load them immediately before decision spans. For bidirectional encoders, distribute anchors around decision spans to enhance robustness. In mixture-of-experts architectures, use sufficient anchor variety to ensure reliable routing to structural experts without over-fragmenting attention.

## Discussion

This analysis reveals that the relationship between anchor properties and attention dynamics involves fundamental trade-offs rather than monotonic improvements. The key insight concerns the distinction between margin effects, which provide unambiguous benefits, and redundancy effects, which require careful optimization due to competing fidelity and entropy considerations.

Future work should empirically validate these theoretical predictions across different model architectures and rule domains. The fidelity-entropy trade-off deserves further investigation in settings with heterogeneous anchor types and complex rule structures. Extensions to multi-modal and retrieval-augmented systems present additional opportunities for theoretical development.

## Conclusion

Ensuring that large language models behave safely and lawfully requires principled approaches to rule encoding that account for information-theoretic properties of attention mechanisms. Our analysis demonstrates that low-entropy, anchor-rich encodings can reduce attention entropy and improve rule retrieval, but reveals previously unrecognized trade-offs between anchor redundancy and entropy that must be carefully managed.

The formal framework establishes bounds on pointer fidelity across multiple attention architectures and proves that hot reloading of verified rule sets, combined with external symbolic verification, provably increases compliance probability. These findings underscore the necessity of dynamic rule management and dual enforcement mechanisms to protect LLM-based agents against prompt injection attacks while maintaining compliance in evolving professional domains.

Understanding these trade-offs provides a more principled foundation for designing rule-encoding strategies in large language model applications, emphasizing the importance of anchor distinctiveness and strategic placement over simple duplication approaches.